\documentclass[letterpaper, 10 pt, conference]{ieeeconf}
\IEEEoverridecommandlockouts

\usepackage{amsmath,amsfonts,amssymb,mathrsfs}
\usepackage{graphicx}
\usepackage{psfrag,graphicx,epsfig,epsf}
\usepackage[ruled,linesnumbered, noend]{algorithm2e}
\usepackage{fancyhdr}
\usepackage{wrapfig}
\usepackage{mathtools}
\usepackage[permil]{overpic}
\usepackage{url}
\usepackage{color}
\usepackage{verbatim}
\usepackage{xspace}
\usepackage{cite} 
\usepackage[export]{adjustbox}
\usepackage{graphicx}
\usepackage{multirow}

\usepackage{caption}
\usepackage{subcaption}

\usepackage{url}
\usepackage{breakurl}
\usepackage[breaklinks]{hyperref}

\makeatletter
\newcommand{\removelatexerror}{\let\@latex@error\@gobble}
\makeatother

\title{\LARGE \bf Perspectives on Sim2Real Transfer for Robotics:\\A Summary of the R:SS 2020 Workshop}

\author{Sebastian Höfer$^{*,1}$, Kostas Bekris$^{1,2}$, Ankur Handa$^3$, Juan Camilo Gamboa$^{4}$, Florian Golemo$^{5,6}$, \\
Melissa Mozifian$^{4}$,
Chris Atkeson$^{7}$, Dieter Fox$^{3,8}$, Ken Goldberg$^{9}$, John Leonard$^{10}$, C. Karen Liu$^{11}$, \\
Jan Peters$^{12,13}$, Shuran Song$^{14}$, Peter Welinder$^{15}$, Martha White$^{16}$
\thanks{
$^{*}$Corresponding author: \href{mailto:mail@sebastianhoefer.de}{\texttt{mail@sebastianhoefer.de}}
}
\thanks{
$^1$Amazon Robotics AI
$^2$Rutgers University
$^3$NVIDIA Robotics
$^4$McGill University
$^5$Mila
$^6$ElementAI
${^7}$CMU
${^8}$University of Washington 
${^9}$UC Berkeley
$^{10}$MIT
$^{11}$Georgia Tech
$^{12}$Technische Universität Darmstadt
$^{13}$MPI for Intelligent Systems
$^{14}$Columbia University
$^{15}$OpenAI
$^{16}$University of Alberta
}
}

\begin{document}

\maketitle
\thispagestyle{empty}
\pagestyle{empty}

\begin{abstract}



This report presents the debates, posters, and discussions of the Sim2Real workshop held in conjunction with the 2020 edition of the ``Robotics: Science and System" conference. Twelve leaders of the field took competing debate positions on the definition, viability, and importance of transferring skills from simulation to the real world in the context of robotics problems. The debaters also joined a large panel discussion, answering audience questions and outlining the future of Sim2Real in robotics. Furthermore, we invited extended abstracts to this workshop which are summarized in this report. Based on the workshop, this report concludes with directions for practitioners exploiting this technology and for researchers further exploring open problems in this area.

\end{abstract}

\setcounter{footnote}{16}

\section{Introduction}

Simulation is an important tool for developing robotic agents. While simulation has been well-established for robotics education and integrated robot software testing for years, there is an ongoing debate in the research community about the ability to transfer robotics skills learned in simulation to reality, a concept termed \emph{Sim2Real transfer}.

\begin{figure*}
	\centerline{
		\includegraphics[width=0.5\linewidth]{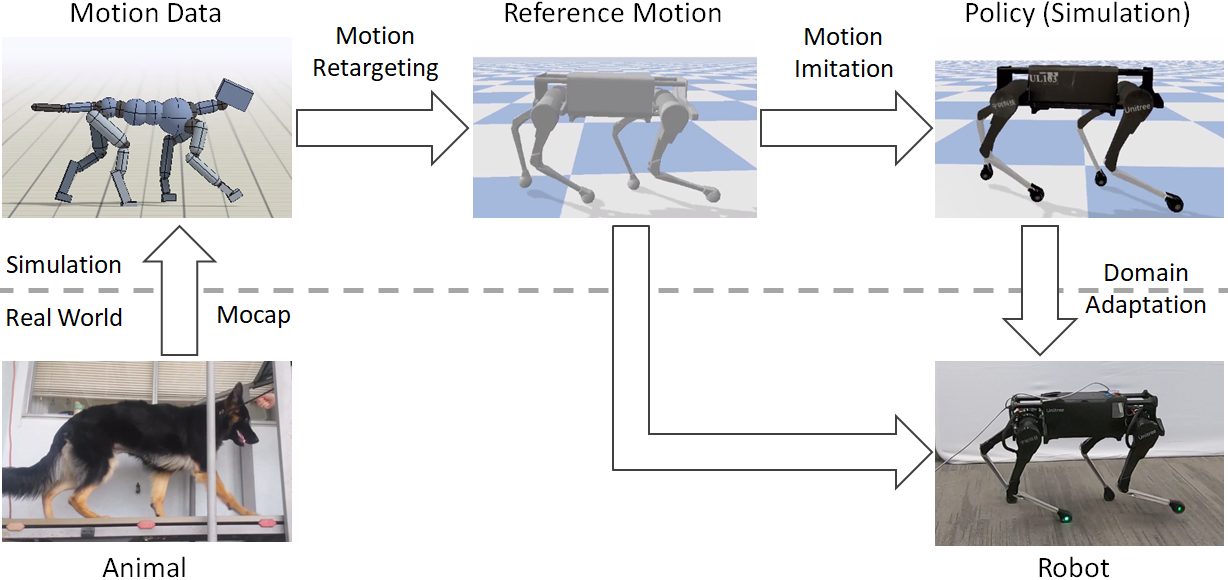}
		\includegraphics[height=1.2in]{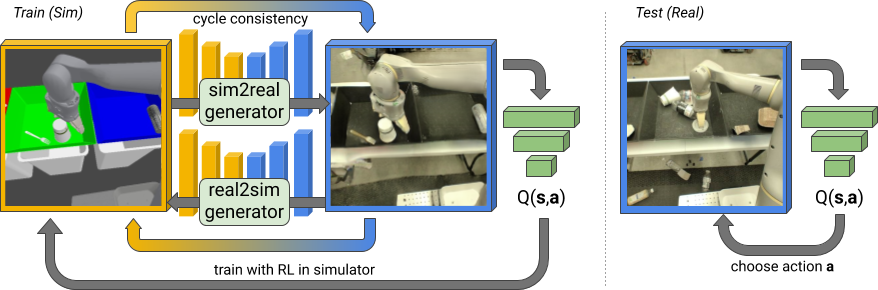}
	}
	\centerline{
		\hfill
		\makebox[0.50\linewidth][c] { \footnotesize{(a) Learning locomotion skills by imitating animals \cite{Peng:etal2020} }}
		\hfill
		\makebox[0.50\linewidth][c] { \footnotesize{(b) RL-CycleGAN \cite{Rao:etal2020:RLCyclegan}}}
	}
	\vspace{5mm}
	\centerline{
		\includegraphics[width=\linewidth]{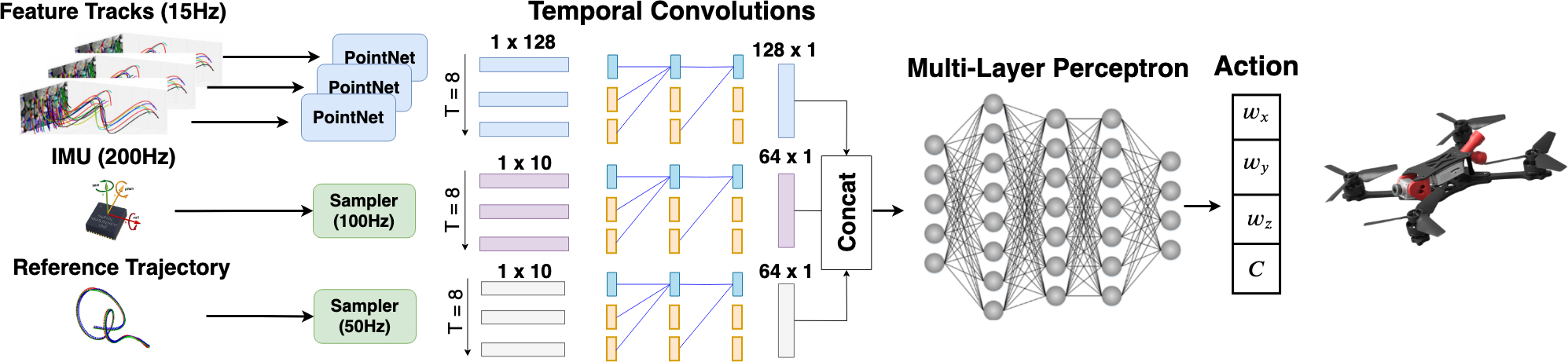}
	}
	\centerline{
		\hfill
		\makebox[\linewidth][c] { \footnotesize{(c) Deep Drone Acrobatics \cite{Kaufmann:etal:2020RSS}}}
	}
	\vspace{5mm}
	\centerline{
		\includegraphics[width=0.5\linewidth]{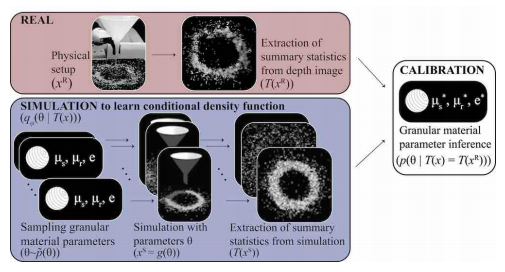}
		\includegraphics[height=1.4in]{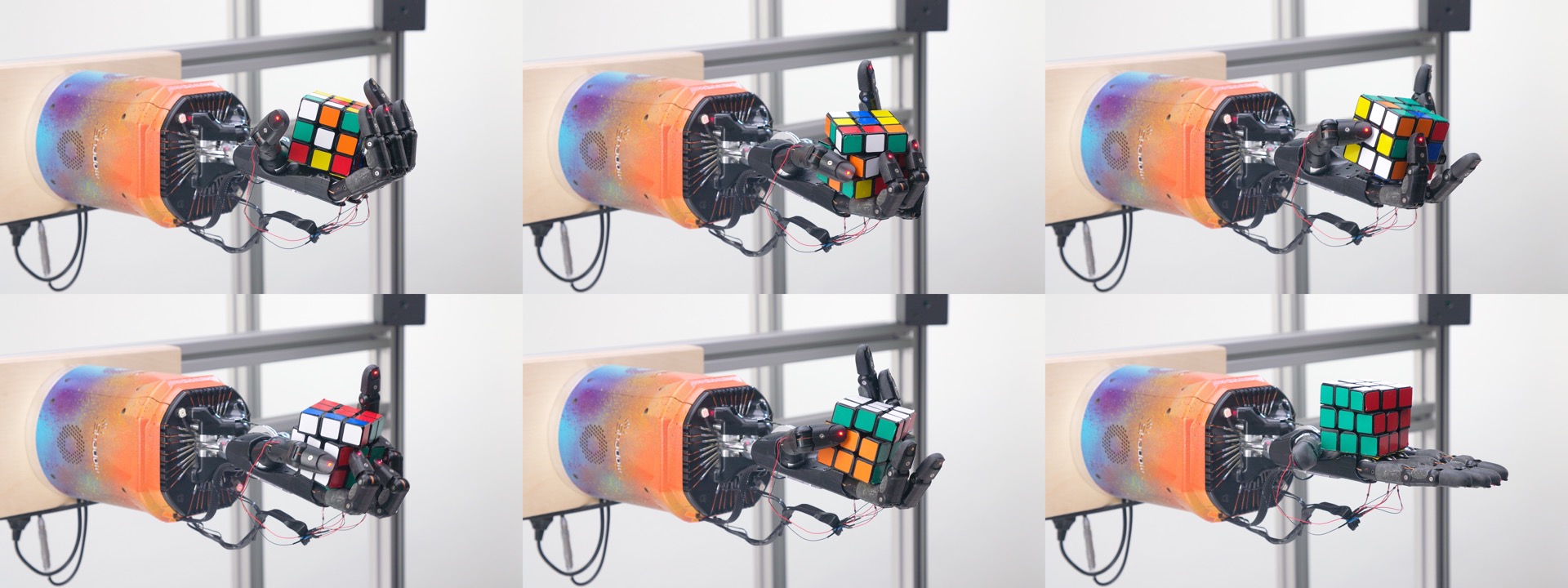}
	}
	\centerline{
		\hfill
		\makebox[0.50\linewidth][c] { \footnotesize{(d) Sim to real on granular media \cite{Matl:etal:ICRA2020} }}
		\hfill
		\makebox[0.50\linewidth][c] { \footnotesize{(e) Manipulating Rubik's cube with dexterous hand \cite{openai:etal:2019}}}
	}

	\caption{\small
		Various state-of-the-art applications of Sim2Real in robotics published in 2019-2020.} \vspace{1mm}
	\label{fig:sim-to-real-2020}
\end{figure*}

The appeal of learning in simulation stems from the fact that it can be faster than real-time, cheaper, safer, and more informative (\textit{e.g.} providing perfect ground truth labels) than real-world experimentation. Recent work in Sim2Real has studied difficult real-world robotic problems related to autonomous driving, grasping, or in-hand manipulation with policies trained in simulation only. Fig. \ref{fig:sim-to-real-2020} highlights some of the work carried out this year. However, Sim2Real still faces significant challenges, and it remains an open question to which extent and in which problem domains Sim2Real can compete with or outperform techniques based on real-world experimentation.

To shed light on these questions, we organized a scientific workshop on the topic of Sim2Real transfer, which was held at the \emph{Robotics: Science and Systems (R:SS) 2020}\footnote{All debate videos and abstracts have been made available at\\ \url{https://sim2real.github.io}} conference. We invited subject matter experts to debate the state of the art and future of the field. The highlight of the workshop were three debates with the following topics and corresponding controversial key statements:

\begin{itemize}
	\item \textbf{\emph{Why} should we invest in Sim2Real?} - ``Sim2Real is a waste of time and money.''
	\item \textbf{\emph{What} is Sim2Real? }- ``Sim2Real is old news. It is just [model-based reinforcement learning $\vert$ domain randomization $\vert$ system identification $\vert$ $\ldots$].”
	\item \textbf{\emph{How} should we apply Sim2Real?} - ``For successful sim2real transfer, there is no alternative to accurate simulation.“ 
\end{itemize}

In order to ensure that a wide range of arguments were considered during the debates, we divided the debate participants equally into proponents and opponents (two debaters for each side in each debate) and asked them to adhere to their assigned role throughout the debate.

The workshop also featured contributed work in the form of 18 peer-reviewed two-page abstracts and was concluded with a panel discussion including all debate participants.


The goal of this article is to summarize the workshop and draw conclusions about Sim2Real for robotics from both from a practitioner’s as well as a researcher’s perspective. In the first part, we summarize the three debates and the concluding panel discussion, highlighting the main arguments brought forward during the debates\footnote{In general, we avoid highlighting the debaters' names since they were assigned fixed positions which might not reflect their actual beliefs.} and the panel. The second part gives an overview of the contributed work, highlighting the main themes and summarizing individual contributions. In the last section, we draw lessons from the workshop to give concrete recommendations to practitioners on how to apply Sim2Real to their robotic applications, and to gather fundamental open questions in Sim2Real that require the attention of future research.

\section{The Sim2Real Debates}

\subsection{Debate 1: Why should we invest in Sim2Real?}

The debates started off by questioning the motivation for using Sim2Real transfer. Chris Atkeson and Abhinav Gupta argued in favor of and Ken Goldberg and Peter Welinder against the controversial statement \emph{"Sim2Real is waste of time and money"}. We present the key topics covered in the debate which can be broadly clustered into the actual monetary cost and the effectiveness/time saved by Sim2Real. 

\textbf{Sim2Real is cheap.}
The opponent side argued that Sim2Real is significantly cheaper than real world experimentation due to the reduced cost in building and maintaining real robots. The proponents countered that Sim2Real being cheap is a myth as training a policy for OpenAI's Rubik's cube costed several thousands of dollars for simulation alone and over a year of development was necessary to successfully transfer policies to the real world~\cite{Openai2019:Solving}. The opponents argued that the high cost reflects the fact that Sim2Real was applied to a problem of such complexity for the first time, and that its cost will reduce significantly, similar to how the cost of robotic hardware decreased over the last decades.

\textbf{Sim2Real democratizes research vs. ``Sim2Null".} Given the low cost of simulation compared to real-robot experiments, simulation clearly lowers the entry barrier for students and researchers. However, there was consensus that easy-to-use simulation environments like OpenAI gym also lead to a rise of simulation-only research that does not generalize to the real world -- a concept Ken Goldberg termed \emph{Sim2Null}.

\textbf{Sim2Real is diverse.}
Another variant of the cost argument relates to the diversity of data that can be acquired through simulation. While it is possible to sample millions of variants of the same task using techniques like domain randomization, the proponent side suggested that data collection in the real world can generate much more diverse data sets cheaply, for example by renting apartments and exploring them with robots. All agreed that realistic non-deterministic simulation is understudied.

\textbf{Sim2Real already works.}
The opponents highlighted that there are various successful applications of Sim2Real, such as in civil engineering and aircraft design, but also in robotic grasping \cite{mahler_learning_2019}. The proponents objected that "Sim2Real works" is not a well-defined statement, and that many approaches, in particular policies trained in OpenAI gym, generalize poorly if at all to the real world.

\textbf{Simulation as a necessary but not sufficient condition for success in real}. Another argument revolved around the question whether successful performance in simulation is a necessary condition for success in the real world. The opponent side argued for this view, highlighting the opportunity to accelerate research using simulation. The proponents questioned the idea of simulation as a necessary condition, since overly simplified modeling of the world and its modalities may render the simulated problem much harder than the real one.

\textbf{Sim2Real is safe.} There was agreement that experimentation in Sim2Real is inherently safe and thus useful for dangerous tasks and exploration. However, the proponent side argued that robots are becoming safer themselves, for example soft robots.

\textbf{Sim2Gradstudent2Real.} All panelists agreed that the state of the art in Sim2Real involves a large amount of manual tuning. The proponent side criticized this fact and suggested that useful Sim2Real research should be concerned with rigorous modeling and problem understanding \cite{hwangbo_learning_2019}. In contrast, the opponent side argued that this situation motivates investigating in data-efficient general-purpose Sim2Real methods.

In the closing remarks, all speakers unanimously agreed that simulations alone are not sufficient. There was agreement that simulation tasks and benchmarks need to become more challenging and that researchers need to run convincing real-world experiments rather than drawing general conclusions from simulation-only experiments.

\subsection{Debate 2: What Is Sim2Real?}

The second debate topic addressed the question of whether Sim2Real qualifies as a methodology or field of its own. Jan Peters and Martha White argued in favor of and Greg Dudek and John Leonard against the controversial statement \emph{"Sim2Real is old news. It is just [model-based Reinforcement Learning $\vert$ domain randomization $\vert$ system identification $\vert$ $\ldots$]"}.
The debate centered around the issues of building upon existing work, providing grounds for a community to form, and the specifics of the field that make it unique.

\textbf{Sim2Real is prior art}. The proponent side's main argument was that current Sim2Real methods date back to the 1960s to 1980s, with the main difference being the scale of computational resources available. 
For example, Real2Sim can be regarded as system identification and Sim2Real as adaptive control: Both combine analytical models with parametric ones (such as shallow neural networks) in order to model complex systems and learn controllers. Similarly, model-based reinforcement learning can be regarded as a method that iteratively refines a poor model or simulation of the world to learn a controller for the real world. 

\textbf{Sim2Real brings together distant research communities.} While acknowledging that the techniques used in Sim2Real are not necessarily new, the opposing side argued that Sim2Real approaches have had significant impact on a wide variety of research domains beyond control (like perception and physical modelling). Therefore, treating Sim2Real as a field of its own provides value by connecting researchers from different disciplines, including robotics, computer vision, control theory, physics-based modelling, reinforcement learning and simulation-based inference. 


\textbf{Impact of the reality gap on Sim2Real}. Both sides agreed that highly accurate simulations of reality remain a distant “pipe dream” for robotics: there will always exist a reality gap, for any level of sophistication of computer simulations. The debaters disagreed on the impact of this statement on the debate topic. The opponent side argued that the inability to close the reality gap justifies the existence of Sim2Real as a field of its own: Rather than reducing Sim2Real to making simulators more accurate, Sim2Real would then be the field that studies how useful information can be extracted from models that are not entirely accurate. The proponents countered that if Sim2Real is not about closing the reality gap with better modelling, it would simply re-brand existing techniques in robust control.




\subsection{Debate 3: How should we apply Sim2Real?}

The goal of the third debate was to shed light on the relationship between Sim2Real methods and accurate simulations. Dieter Fox and Karen Liu argued in favor and Anca Dragan and Shuran Song against the controversial statement \emph{"For successful Sim2Real transfer, there is no alternative to accurate simulation"}. The debate touched upon a variety of topics, from levels of abstraction to the trade-off between accuracy and generalization. 


\textbf{Right level of abstraction.} The debaters agreed that finding the right level of abstraction is key, yet difficult as it presupposes knowing what is relevant to the task. Both sides agreed that current simulators do not model physical phenomena sufficiently well. While the proponent side concluded that this situation calls for investing into better modeling at lower levels of abstraction, the opponent side argued that high-level abstraction suffices, comparing simulation of physical interaction with human behavior: Robots need to learn to collaborate with humans without simulating the entire human brain, so why would physical interaction require low-level simulation of physics? Furthermore, humans themselves learn from ``abstract simulation'', such as books and movies. 

\textbf{Imprecision.} Assuming a given level of abstraction, the debaters discussed the implications of imprecise simulation. The opponents argued that accurate modelling is expensive and often infeasible. Additionally, any abstraction -- and thus, any simulator -- inevitably results in a loss of accuracy due to the fact that it does not capture the underlying process.
The proponent side argued that imprecise simulators at any level of abstraction are useless because, once deployed to the real world, the agent has to unlearn a lot of what it had learned in the imprecise simulator. The only way to get around this issue is heavy domain randomization over a large set of simulation parameters, which however does not scale for more complex tasks and environments. The opposing side argued that exactly this imperfection and deliberate bias causes model resilience and policy generalization.

\textbf{Generalization and Extrapolation.} Another argument of the proponents against imprecise simulation concerned its limited generalization capability: A precise simulator is a useful building block to solve many different problems, while an imprecise simulator will work for some but not other tasks. The proponent side explicitly distinguished general-purpose simulators from  task-specific internal models, which do not need to be precise but only serve a single purpose.

\textbf{Uncertainty and Noise.} The speakers agreed upon the importance of explicit uncertainty estimates, \emph{i.e.} modeling where the simulator is uncertain about its future state predictions. Such estimates, while difficult to obtain, would be useful to identify failure cases early on or leverage them to collect data for improving the simulator.

\section{Panel Discussion}

The workshop was concluded with a panel discussion. During the panel, the debaters were allowed to express their personal opinions and a few debaters noticed that their opinions had reshaped after defending extreme positions during the debates. The lively panel was a free-form discussion with some recurring topics which we summarize in this section.

{\bf Task-agnostic vs. task-specific simulation.}
An important question to the panel was whether simulators should be task-agnostic or task-specific.
The panelists agreed that there are diminishing returns in improving task-agnostic simulation as we ultimately care about task performance, not simulation accuracy. Moreover, certain domains such as underwater robotics or scenarios involving human interaction are still beyond the scope of state-of-the-art simulators. At the same time, task-agnostic simulators are regarded as the only means to provide tools that generalize across a sufficiently large range of problems. The panelists mentioned \emph{domain randomization} as a success story that shows how task-agnostic state-of-the-art simulation tools allow to learn real-world policies. At the same time, there was agreement that it remains an open question how to apply this technique at scale.

Several alternatives to task-agnostic simulation were considered. One proposition was to invest in \emph{task-agnostic but customizable} simulators. Such a simulator could be obtained by closing the loop and iteratively tuning the simulation to the world using Real2Sim and system identification.

The panelists identified \emph{differentiable simulators} as another promising research direction. The main benefit of this direction is that it moves simulation closer to models which can be used in model-based optimization, allowing for computing derivatives and thus improving the controller directly. There were doubts whether full differentiability is needed since many real world phenomena are not differentiable and this would inevitably make simulator design more complex.

\emph{Meta learning} was identified as an promising orthogonal direction. The key idea is to learn to adapt to a changing world (in simulation) rather than attempting to learn all the characteristics of the world.


{\bf Sim2Real as a research discipline on its own (with its own symposia and academic curriculum).}  
The panelists acknowledged the importance of Sim2Real in its own right, hypothesizing that Sim2Real could be a way of building bridges between disciplines across non-traditional paths, potentially through a Sim2Real grand challenge. The panel agreed that we should refrain from introducing new symposia and conferences to avoid further fragmentation of the robotics community, and to continue exploring Sim2Real in scientific workshops.

{\bf The value of robotics research performed exclusively in simulation (Sim2Sim).} There was wide agreement that the ultimate goal is to design robotic systems that live in the real-world. Nevertheless, simulation-only results can provide valuable insights, for examples by providing repeatable benchmarks on standardized tasks.
Therefore, there should not be a rejection by default of papers that do not contain real-world experiments. At the same time, authors need to do due diligence and provide convincing evidence about the real-world applicability of new approaches that are only validated in simulation. Moreover, the community should invest into creating simulators that simplify the job of researchers as well as practitioners and allow for better Sim2Real transfer. 

\section{Presented Abstracts}

18 short research abstracts were selected to be presented in the workshop during three 20-minutes interactive sessions. Authors were invited to submit work on topics related to Sim2Real, including methods for improving simulator accuracy, training robust controllers, formalizing the Sim2Real problem, assessing the difficulty of and benchmarking strategies for Sim2Real. Below, we provide a short summary of each work.

\subsection{On the formalization and difficulty of Sim2Real}
One of the standing issues in Sim2Real is the lack of a framework for comparing Sim2Real solutions. Paull \emph{et al.}~\cite{Paull:2020aa} propose to distinguish between the use of the simulator as a predictor vs. as a teacher, and argue that regardless of its role, the value of any simulator is its impact on final task performance.

On the topic of the difficulty of Sim2Real, multiple submissions explored the question whether current simulators are adequate for solving complex manipulation and locomotion tasks. Zhang~\cite{Zhang:2020aa} argues that the gap between current simulators for contact-rich tasks is too wide to expect any positive transfer, even when considering domain randomization. Rizzardo~\emph{et al.}~\cite{Rizzardo:2020aa} describe one such contact-rich task, precision agriculture, where simulation of fruit picking is a bottleneck for Sim2Real. Dao~\emph{et al.}~\cite{Dao:2020aa} show show how a state-of-the-art simulator, with its limitations in simulating contacts, can be used to train controllers that successfully transfer to real a walking bipedal robot without using any domain randomization.



Finally, on the topic of benchmarking, Kadian \emph{et al.}~\cite{Kadian:2020aa} introduce Habitat-PyRobot Bridge (HaPy), a library for seamless execution of code on simulated agents and robots, and propose using the Sim-vs-Real Correlation Coefficient (SRCC) as a transfer metric: to evaluate whether improvements in performance in simulation are correlated with improvements in the real world.


\subsection{On bridging the reality gap with existing simulators}
A successful technique for Sim2Real is to explicitly design or learn intermediate representations that generalize between simulation data and real data. Zhang \emph{et al.}\cite{Zhang:2020ab} propose to using an off-the-shelf 3D object detector to train obstacle avoidance policies for manipulation that directly transfer to the real world without the need of a visually accurate simulator. Related to this approach, Liang \emph{et al.}\cite{Liang:2020aa} train obstacle avoidance policies for mobile robot navigation in simulated environments to navigate crowds. Here, the intermediate representation is provided by laser scan readings, which are easier to simulate than cameras. Yan~\emph{et al.}~\cite{Yan:2020aa} propose to learn intermediate representations for deformable object manipulation by jointly optimizing both the visual representation model and the dynamics model using contrastive estimation. Antonova~\emph{et al.}~\cite{Antonova:2020ab} propose a meta-representation learning approach that infers analytic relations from simulation and leverages these relations when learning intermediate representations in the real domain.

A related approach for vision tasks is to use off-the-shelf photorealistic rendering engines to produce self-labeled data. 
Denninger~\emph{et al.}~\cite{Denninger:2020aa} introduce \textit{BlenderProc}, a pipeline for procedural generation of labelled data for vision tasks, which the authors demonstrate on the task of image segmentation.  Wu~\emph{et al.}~\cite{Wu:2020aa} leverage domain randomization to generalize deformable object manipulation policies to the real domain.

Finally, the Sim2Real gap can be addressed at the policy level. Malmir \emph{et al.}\cite{Malmir:2020aa} introduce a method that leverages the discrepancies between inverse dynamics in simulation and the real world to compute corrections to controllers learned in simulation. Josifovski \emph{et al.}~\cite{Josifovski:2020aa} propose reducing the Sim2Real gap using curriculum learning, by training the robot on increasingly more difficult versions of the task.





\subsection{On bridging the reality gap with improved models and real-world data}
Based on the intuition that more accurate models are more likely to transfer to real, several submissions propose leveraging real-world data for improving simulators. 

Raparthy~\emph{et al.}~\cite{Raparthy:2020aa} propose a curiosity-based data collection scheme  for training a physics-based simulator with a residual model for the discrepancies between real-world and simulated data. Heiden~\emph{et al.}~\cite{Heiden:2020aa} propose a differentiable formulation of rigid-body dynamics that allows training both a residual physics model along with the internal parameters of the simulator, allowing to backpropagate through the physics-based model.


Possas \emph{et al.}~\cite{Possas:2020aa} take an alternative approach, where the simulator is treated as a black-box, and real-world data is used to infer the parameters of the simulator online as a robot is learning to execute a task. Antonova \emph{et al.}~\cite{Antonova:2020aa} propose a related method, where inference is done with Gaussian Processes with kernels learned from simulation data.
Similarly, Desai \emph{et al.}~\cite{Desai:2020aa} propose to reduce the distribution mismatch between  simulator and real world by learning behaviors that mimic the observations of behavior demonstrations.



\section{Key Directions Forward}


The workshop debates and contributed papers provide a view on the state-of-the-art in Sim2Real techniques and open problems in this area. This section summarizes the findings and conclusions from the workshop, which we classify as belonging to either a practitioner's or a researcher's perspective. For the practitioner's perspective, we aim to give concrete advice on applying existing Sim2Real techniques, their capabilities and their limitations\footnote{We mainly focus on established techniques that have relevance for the workshop. For a recent, more complete survey see e.g. \cite{zhao2020simtoreal}.}. For the researcher's perspective, we state fundamental research problems identified during the workshop, hoping to inspire future research endeavors. 


\subsection{Practitioner’s View: What Can Sim2Real Already Achieve?}

The first question that practitioners should ask themselves is whether their problem setting would benefit from Sim2Real techniques. The following list of requirements constitute criteria that can provide guidance in answering this question:
\begin{itemize}
\item {\bf Bootstrapping:} Is it possible to obtain a proof of concepts relevant to the real-world problem in simulation? Simulators can be extremely helpful for scoping the problem and initial hypothesis testing. 
For learning and adaptive systems, learning in a carefully tuned simulator is an effective way to bootstrap learning in the real world.
\item {\bf Long-term data starvation:} Is the effort in collecting real world data, especially labeled data, considerably higher, more dangerous or even infeasible relative to the effort of developing a simulator? Simulators can be a valuable and endless source of labeled synthetic data, in particular if data is and remains hard to collect due to cost or safety constraints. However, real data should be preferred if available in sufficient quantities.
\item {\bf Hardware-in-the-loop optimization:} Is the goal to co-design hardware and software? In such scenarios there may be no alternative to simulation. Sim2Real techniques can be useful for learning and exploring the hardware design space, which may be impractical without a simulator. 

\item {\bf Common computer vision problems:} Is the problem mainly a computer vision task? Object detection, image segmentation and geometry estimation have received significant attention in the last decade, with the role of synthetic data becoming more prominent as part of their solutions.
\end{itemize}

To address the requirements from the previous list, we summarize some of the {\bf techniques} that have been shown to work well in practice.


{\bf Domain randomization (DR)} is a simple yet powerful technique to augment the training set, and particularly well suited for training deep neural networks. The main challenges, however, correspond to finding the right set of parameters to randomize. The cost of applying DR grows exponentially with the number of parameters.

{\bf Explicit transferable abstractions} A powerful alternative to DR is identifying explicit features or state abstractions, which transfer easily between simulated and real data. For example, pre-processing images into segmented images, edges, depth maps or using off-the-shelf image-to-image translation methods circumvents the problem of generating realistic images in simulation, but it still requires careful tuning. 

{\bf Combining analytical modeling with system identification (SysID)} In some settings we can identify challenging problem aspects, which can be modeled analytically and apply SysID/ML for them. This approach has been successfully used for grasping and locomotion, and is commonplace in model-based control. Nevertheless, finding appropriate models is a challenging task. 

{\bf Meta-Learning and Curriculum Learning:} A common limitation of Sim2Real is its application to situations where the robot dynamics are hard to simulate, \emph{e.g.} simulating the transformation from control signals to motor torques. While learning SysID modules is a possible way of exploiting simulated data, meta-learning provides an alternative for directly learning systems that adapt to changing environments or erroneous models. DR techniques can be used to provide a task distribution for meta-learning. Furthermore, controlling DR simulators can be an effective way of controlling the difficulty of learning. 



{\bf Depth \& RGB rendering} 
Depth sensors including their noise models can be effectively simulated and transfer smoothly to the real world 
RGB rendering is also catching up, and recent work has shown increasingly better Sim2Real generalization~\cite{mahler_learning_2019}. 
These approaches, however, generally require high-quality CAD models to be available.


{\bf Domain adaption via GANs} A complementary technique is to enhance the quality of synthetically generated images using domain adaption techniques~\cite{zhu2020unpaired}. One challenge here is to make sure the labels are not distorted.

As a final remark, it is important to be aware of the fact that Sim2Real techniques are not yet Plug-\&-Play nor completely automated. Their application still requires careful human attention for understanding the problem and for tuning the parameters of the chosen approach.

\subsection{Researcher’s View: What Are the Open Questions?}

While the current Sim2Real methods have shown promising results for real world problems, there remain important open questions and unsolved problems. 

{\bf Systematic and Efficient Domain Randomization (DR)} Existing DR methods are computationally wasteful, even after careful selection of the randomization distributions. Furthermore, selecting the appropriate randomization distribution may be as hard as the problems we would like to address with DR. What methods can help us automate the selection of a DR distribution? What are the data requirements of these methods? Can we establish how hard finding a DR distribution is? Can we accelerate learning using a DR schedule or curriculum?


{\bf Contact-Rich Tasks} Many simulators struggle to simulate physically realistic contacts. Contacts with friction is in general an NP-hard problem and many simulators implement contacts with isotropic Coulomb friction, which is an inaccurate approximation of real contacts. What level of accuracy is necessary for applying Sim2Real techniques to these important aspects for manipulation and locomotion?

{\bf Differentiable Simulators} There is growing interest in creating simulation pipelines, which allows differentiating the stack of operations involved in simulating dynamics and rendering. This is a promising direction as that could allow rapid testing and learning of the simulation parameters. Is differentiability a requirement for successful Sim2Real? Do the benefits of differentiable simulation outweigh the cost of replacing existing black-box simulators? Are there fundamental limits to closing the reality gap with differentiable simulation?

{\bf Highly Accurate Simulations} As time goes on, the capabilities for high-fidelity simulation and rendering keep improving in accuracy and speed. Will this trend continue to a point where high fidelity simulation becomes an easily accessible commodity for most robotics tasks? How can we improve the simulation accuracy and speed for fabrics, deformable materials as well as contact-rich tasks? Are there any fundamental limits to simulation?


{\bf Production-level Sim2Real} How can we push task performance using Sim2Real to production level (99.99\% success rates)?





\section{Conclusion}

The workshop highlighted the fact that Sim2Real is a “hot topic”, with many academic and industrial institutions putting significant resources into  investigating the applying and developing Sim2Real approaches. While there was no agreement in the debates as to how far Sim2Real will take us in the future and whether it is the right approach in general, there was consensus on what Sim2Real can do for robotics applications today. A community is forming around the problem with a steadily increasing number of submissions. This may require the introduction of this keyword as a separate category during submission to the main robotics venues, if not potentially devoted meetings in this area. While the future will tell how Sim2Real will evolve and what impact it may have on the next breakthroughs in robotics, it is an exciting area to explore from many different perspectives. 

\section*{Acknowledgements}
The authors would like to thank Anca Dragan, Gregory Dudek and Abhinav Gupta for their contributions to the debates and panel discussion as well as the R:SS 2020 organizers for hosting our workshop.
{
\bibliographystyle{plain}
\bibliography{main}
}

\end{document}